\documentclass[a4paper]{article}

\usepackage{INTERSPEECH2022}
\usepackage{amsmath,graphicx,blindtext,color,hyperref,tikz,tabularx,verbatim}
\usetikzlibrary{matrix,chains,positioning,decorations.pathreplacing,arrows,arrows.meta,external,shapes, shapes.geometric,shapes.misc,calc,external}

\DeclareMathOperator*{\argmax}{arg\,max}

\DeclareMathOperator*{\shortestpath}{shortest\_path}

\newcommand{\figref}[1]{Figure~\ref{#1}}

\title{Deciphering Speech: a Zero-Resource Approach to \\Cross-Lingual Transfer in ASR}
\name{Ond\v{r}ej Klejch, Electra Wallington, Peter Bell\thanks{This work was supported by EPSRC Project EP/T024976/1 (Unmute).}}
\address{Centre for Speech Technology Research, University of Edinburgh, United Kingdom
\email{\{o.klejch, electra.wallington, peter.bell\}@ed.ac.uk}}

\begin{document}

\maketitle
\begin{abstract}
We present a method for cross-lingual training an ASR system using absolutely no transcribed training data from the target language, and with no phonetic knowledge of the language in question.  Our approach uses a novel application of a decipherment algorithm, which operates given only unpaired speech and text data from the target language.  We apply this decipherment to phone sequences generated by a universal phone recogniser trained on out-of-language speech corpora, which we follow with flat-start semi-supervised training to obtain an acoustic model for the new language.  To the best of our knowledge, this is the first practical approach to zero-resource cross-lingual ASR which does not rely on any hand-crafted phonetic information. We carry out experiments on read speech from the GlobalPhone corpus, and show that it is possible to learn a decipherment model on just 20 minutes of data from the target language. When used to generate pseudo-labels for semi-supervised training, we obtain WERs that range from  32.5\% to just 1.9\% absolute worse than the equivalent fully supervised models trained on the same data.
\end{abstract}


\section{Introduction}
\label{sec:intro}

In recent years there has been considerable research devoted to reducing the amount of human effort required to build an automatic speech recognition (ASR) system for a new language.  Conventional ASR training requires large quantities of manually-transcribed training data, as well as a hand-crafted pronunciation dictionary.  Recent grapheme-based hybrid-HMM approaches~\cite{liu2020multilingual} have shown success at removing the need for explicit pronunciation knowledge, whilst more recent end-to-end systems~\cite{chan2016listen} have removed the need for a lexicon entirely by modelling output tokens at the character or word-piece level.  However, transcribed training data is typically still required, with end-to-end systems being particularly data hungry.  

The process of manual transcription can be extremely time-consuming and expensive.  Consequently a body of research has focused on reducing the need for such data, for example through the use of approximately-matching ``in-the-wild'' speech and text data, known as lightly-supervised training~\cite{lamel2002lightly}, and through the use of unlabelled data transcribed with a seed model, known as semi-supervised training~\cite{lamel2002unsupervised}.  However, in both cases, manual expertise is required to train the initial model.

In his 2012 position paper, Glass \cite{glass2012towards} described the road towards unsupervised speech processing through a set of scenarios that, he noted, ``might seem increasingly outlandish and impractical''.  He suggested a move from ``expert-based'' systems, with a dictionary and phoneme set provided, through ``data-based'' systems with parallel speech and text data, to what he called ``decipher-based'' systems, through which ASR training could be achieved using entirely untranscribed speech, together with unpaired text data.  This scenario has the significant advantage that for any languages with a significant web presence at least, both resources are likely to be relatively abundant without any human effort.

Since Glass's paper, significant effort has been devoted to this so-called ``zero-resource'' scenario.  Approaches to this problem tend to fall into two categories: those attempting to learn phoneme- or word-like patterns from speech in a bottom up manner, often motivated by child speech learning \cite{kamper17_segmental,hermann21_multi}; and those using cross-lingual information to inform the target model.      
The latter category extends a long strand of research into cross-lingual ASR methods -- which seek to improve supervised training on a target language through the use of out-of-language language data -- to the case where no transcribed data exists for the target language.  There have been a variety of recent approaches to this problem, all of which in one way or another address the problem of matching the modelling units of the out-of-language model to meaningful units in the target language.  The earliest approaches used IPA-based phone mapping schemes \cite{vu2011cross} %
whilst more recent related methods have used automatic multilingual pronunciation mining from the web \cite{lee2020massively} or cross-lingual transfer from languages with similar orthographies \cite{wiesner2019zero}.  \cite{li2021hierarchical} uses knowledge of compositional phonetics in the target language to remove the need for resources in the target language, building on earlier supervised approaches such as \cite{hai2014cross}, whilst \cite{shinozaki2017semi} used a semi-supervised approach, extending an initial lexicon using unpaired phonemic transcripts and text data. 

Separately, there has been significant work in towards building language-universal systems, generally with shared phonetic knowledge \cite{li2020universal}.  These approaches can be problematic due to differing phonotactics between languages \cite{feng2021phonotactics}, though language-specific embeddings may be used~\cite{gao2021zero}.  Again, these methods require knowledge of pronunciations in a target language in order to produce word output.  Purely graphemic multilingual systems have been developed \cite{liu2020multilingual} but require the target language to be in the training set; supervised transliterations approaches have been used in the context of end-to-end systems \cite{khare21_transliteration}.

Purely bottom-up approaches to zero-resource ASR, whilst interesting, have not generally yielded state-of-the-art ASR performance, when compared to cross-lingual methods.  However, groundbreaking work in this area \cite{baevski2021unsupervised} uses Facebook's wav2vec2.0 architecture \cite{baevski2020wav2vec} to produce phone-like sequences in an entirely bottom-up manner, which are then mapped to phonemized text sequences using an adversarial objective \cite{goodfellow14_gans}.  However, this work relies on manually-obtained phone units and a system trained on a large hand-crafted pronunciation dictionary, with the authors noting that it is easier to learn a mapping between the speech audio and phone units.  Further, the wav2vec2.0 models need very large amounts of training data to be effective.

\begin{figure}

\centering

\tikzstyle{atipstyle}=[>={Stealth[inset=0pt,length=6pt,angle'=40,round]}]
\begin{tikzpicture}[atipstyle, node distance=1.25cm, scale=0.8, every node/.style={transform shape}]

\tikzstyle{loss} = [ thick, chamfered rectangle, chamfered rectangle xsep=2cm, text centered, draw=black, minimum width=4cm, minimum height=1cm]
\tikzstyle{data} = [text centered, text width=3.5cm]
\tikzset{linestyle/.style={thick, rounded corners}}
\tikzset{model/.style={rectangle, rounded corners, text centered, inner sep=5pt, text width=3.5cm, minimum width=4cm, minimum height=1cm, draw=black, thick}} 
 
\node (input) [data] {Untranscribed Audio};
\node (upr) [model, above of=input, yshift=0.5cm] {Universal Phone Recogniser};
\node (phones) [data, above of=upr] {Decoded Phones};
\node (decipherment) [model, above of=phones] {Decipherment};
\node (textdata) [data, right of=decipherment, xshift=2.5cm, text width=1.5cm] {Text Data};
\node (pseudolabels) [data, above of=decipherment] {Pseudo Labels};
\node (sst) [model, above of=pseudolabels] {Flat-Start\\Semi-Supervised Training};
\node (seed) [loss, above of=sst, text depth=0.1em, yshift=0.5cm] {Acoustic Model for New Language};

\draw[linestyle]     ($(sst.north west)+(-0.25,0.25)$) rectangle ($(upr.south east)+(0.25,-0.25)$);
\draw [line width=2pt, white] ($(sst.north)+(-0.25,0.25)$) -- ($(sst.north)+(0.25,0.25)$);
\draw [line width=2pt, white] ($(upr.south)+(-0.25,-0.25)$) -- ($(upr.south)+(0.25,-0.25)$);
\draw [line width=2pt, white] ($(decipherment.east)+(0.25,-0.25)$) -- ($(decipherment.east)+(0.25,+0.25)$);

\node (bn_label) [left of=decipherment, rotate=90, yshift=1.25cm] {Zero Resource Cross-Lingual Transfer};

\draw [linestyle, ->,] ($(input.north)$) -- ($(input.north)+(0,0.5)$) -| (upr.south);
\draw [linestyle,] (upr) -- (phones);
\draw [linestyle, ->,] (phones) -- (decipherment);
\draw [linestyle, ->,] (textdata) -- (decipherment);
\draw [linestyle] (decipherment) -- (pseudolabels);
\draw [linestyle, ->] (pseudolabels) -- (sst);
\draw [linestyle, ->] (sst) -- (seed);

\end{tikzpicture}
\caption{A diagram of a zero-resource cross-lingual transfer pipeline.}
\label{fig:pipeline}

\end{figure}
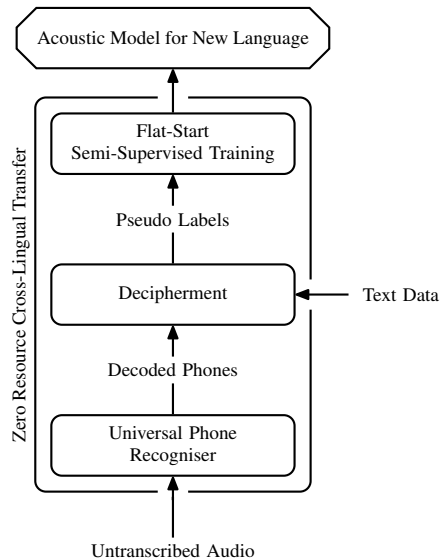

We believe that no prior research has yet achieved Glass's vision of removing both the need for transcribed audio data \textit{and} human phonetic knowledge of the target language, thus building a system with no expert input.  In this paper, we propose to return to his original term of ``decipher-based'' systems.  Inspired by this, and by similar work for unsupervised transliteration and machine translation \cite{ravi2011deciphering,nuhn2019unsupervised}, we here present a method for deciphering speech data using only mismatched text data from the language of interest.  Our method starts with a cross-lingual approach, taking a universal phone recogniser trained on a variety of source languages.  No self-supervised pre-trainined is needed, and
we find that the technique is highly data efficient, requiring just 20 minutes of speech data from the target language to achieve a successful decipherment.  Furthermore, no phonetic knowledge of the target language is used, making the method applicable in principle to almost any language.

\section{Zero-Resource Cross-Lingual Transfer}
\label{sec:cross-lingual_transfer}

Our method for zero-resource cross-lingual transfer uses a three-stage approach.   First a universal phone recogniser transcribes audio into phones. Then, we decipher this phone sequence into graphemes from the target language -- for this, only language models trained on target-language text data are required. Finally, a flat-start semi-supervised training procedure is used to train a new acoustic model using the deciphered pseudo-labels.  The complete pipeline is illustrated in \figref{fig:pipeline}.  We describe the three steps in detail below.

\subsection{Universal Phone Recognition}
The aim of a universal phone recogniser is to phonetically transcribe speech from any language. To achieve good generalisation to unseen languages, it is necessary to train the model on a diverse set of languages, in order to cover as wide a set of phones as possible.  One way to train such a system is to simply pool data and phonemic lexicons and train a multilingual model with a shared phoneme set.  In this work, for simplicity, we use a shared phoneme set to train a conventional hybrid HMM-DNN system on six well-resourced languages.  We are aware that \cite{li2020universal} notes that pooling the phoneme sets is sub-optimal as phonemes might have different surface forms in different languages, and that the use of linguistically-derived allophone mappings \cite{li2020universal} might be beneficial.

\subsection{Decipherment}
\label{sec:decipherment}

The task of decipherment is to convert a cipher into plain natural language, a classic example being deciphering a letter substitution cipher.
The use of this technique to decipher the output of a multilingual phone recogniser is the most significant contribution of this paper.  We start with the work of Knight~\cite{knight1999decoding}, who showed that a noisy-channel framework can be used for decipherment. In this framework the probability of deciphering a cipher X into an English\footnote{In this section we follow the literature in taking English as the target language; of course, in reality we decipher other languages.} text Y is modelled as

\begin{equation}
    \begin{aligned}
       P(Y, X) = P_\text{lex}(X \mid Y) P_\text{lm}(Y),
    \end{aligned}
\end{equation}
where we call $P_\text{lex}(X \mid Y)$ the Lexical model and $P_\text{lm}(Y)$ the language model. The lexical model produces the probability that an English letter $y$ corresponds to a cipher letter $x$ and the Language model $P_\text{lm}(Y)$ assigns probabilities to sequences of English letters. The Language model can be trained on any text corpora and the Lexical model $P_\text{lex}(X \mid Y)$ can be trained in an unsupervised fashion with the Baum Welch algorithm~\cite{baum1970maximization}. Once the lexical model is trained, the most probable English text corresponding to the cipher can be deciphered with the Viterbi algorithm to obtain:
\begin{equation}
    \begin{aligned}
       \hat{Y} = \argmax_Y P_\text{lex}(X \mid Y) P_\text{lm}(Y).
    \end{aligned}
\end{equation}

In the past, decipherment was used in various NLP applications such as unsupervised machine transliteration~\cite{ravi2009learning}, unsupervised machine translation~\cite{ravi2011deciphering,nuhn2014decipherment} or unsupervised Chinese pronunciation learning~\cite{chu2020learning}.
However, phoneme-to-grapheme (P2G) conversion is much more difficult than solving a deterministic substitution cipher for two reasons: first, a grapheme can be mapped to many phonemes, for example English grapheme ``a" can be pronounced as AH, AA, AE or EH. Second, one grapheme can correspond to multiple sequential phonemes, for example ``x" is pronounced as ``K S"; similarly, one phoneme can align with multiple sequential graphemes, for example ``th" is often pronounced as ``DH". Finally, when applying phoneme-to-grapheme conversion at the utterance level, our model needs to be able to perform word segmentation. These challenges are further multiplied when we deal with noisy inputs from the universal phone recogniser.

To be able to deal with insertion and deletions inherent to the P2G, we use the following parameterisation proposed by Nuhn~\cite{nuhn2019unsupervised}:

\begin{equation}
    \begin{aligned}
        \hat{Y} &= \argmax_Y P(Y \mid X) \\
                &= \argmax_Y P_\text{lex}(X \mid Y, A) P_\text{lm}(Y) P_\text{ali}(A) 
    \end{aligned}
\end{equation}
In this parameterisaton the decipherment model consists of three components: Lexical model $P_\text{lex}(X \mid Y)$, Alignment model $P_\text{ali}(A)$, and Language model $P_\text{lm}(Y)$. The random variable $A$ represents a sequence of substitution, insertion and deletion operations.
We can also express decipherment using WFST notation as:
\begin{equation}
    \hat{Y} = \shortestpath (X \circ (L \circ A \circ G)),
\end{equation}
where $X$ is an input phone acceptor, $L$ is the Lexicon model transducer, $A$ is the Alignment model transducer and $G$ is the Language model acceptor.

\begin{figure}
    \centering
    \includegraphics[width=0.6\columnwidth]{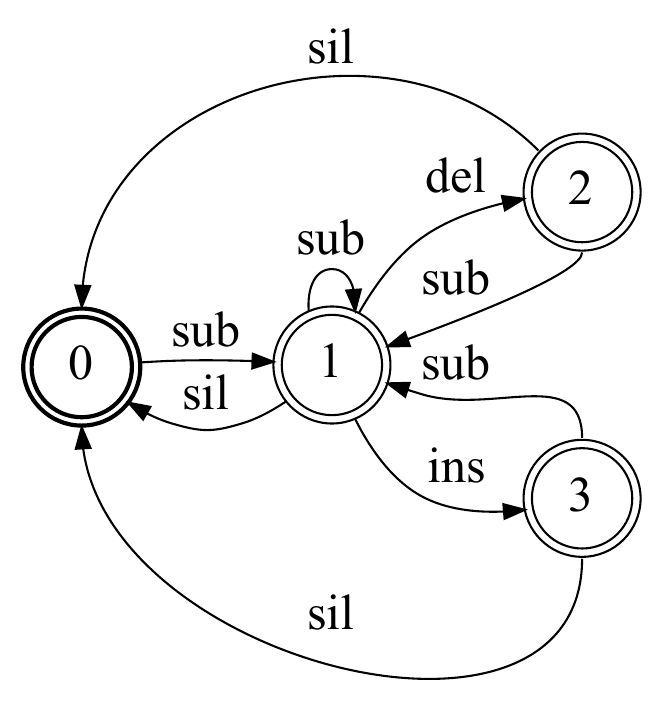}
    \caption{A diagram of an alignment model which allows one insertion or one deletion in a row.}
    \label{fig:alignment-model}
\end{figure}

The role of the \textbf{Lexical model} is to model the probabilities of mapping phones into graphemes, and also to model the probability of phone deletions $P_\text{lex}(x \mid \epsilon)$. The lexical model is implemented as a simple one state flower transducer. It is initialised to allow all possible substitutions but during training the unseen substitutions are pruned from the model, which results in faster training and inference due to a smaller composition. However, since we train on small amounts of data in an unsupervised fashion it is possible that some important arcs are pruned from the model, which can be detrimental. Therefore, following~\cite{nuhn2014decipherment} we smooth the Lexical model at various stages of training with the following equation:
\begin{equation}
    P_\text{lex}^s(x \mid y) = \alpha P_\text{lex}(x \mid y) + \frac{1 - \alpha}{|X|}, 
\end{equation}
where $\alpha$ is a smoothing parameter (we use 0.9) and $|X|$ denotes the size of the input phone-set.
Finally, the lexical model always maps silence phones to silence or a word boundary in the output. This results in faster training/inference -- because silence prunes the space of possible word segmentation -- and more accurate decipherment.

The \textbf{Language model} is the component most important to decipherment, providing information to the training process. The language model predicts the probability of a sequence of graphemes in the target language. Therefore, it is possible to use character n-gram models. It is important to keep in mind that using n-gram models with large contexts results in a big composition when composed with an unpruned lexical model; therefore it is not feasible to use them from the beginning. Hence we start with a simple bigram model and move to using up to 5-gram grapheme models as training progresses. Subsequently, we use a word trigram language model together with a grapheme lexicon for the final round of training. Since the composition of the lexical model, alignment model and the word language model $L \circ A \circ G$ is slow, and is required after every training epoch, we reimplemented the standard composition  $X \circ (L \circ A \circ G)$ with a three-way composition $X \circ (L \circ A) \circ G$~\cite{allauzen2008threeway} . We also use pruning to speed up training and inference with the word language models. Our decipherment pipeline is implemented in OpenFST~\cite{allauzen2007openfst} and its design is heavily inspired by the BaumWelch library from OpenGrm~\cite{roark2012opengrm}.

\subsection{Semi-Supervised Training}

In conventional Semi-Supervised Training (SST) we use a seed-model to create ``pseudo-labels" for untranscribed speech data~\cite{lamel2002unsupervised,manohar2018semi}. In our previous work~\cite{wallington2021learning}, we showed that SST can be successfully even with seed models with WER over 80\% if lattices are used to model uncertainty in the hypotheses. In the previous section we described how decipherment can be used to convert the output of a universal phone recogniser into a sequence of target-language graphemes, to be used as pseudo-labels for untranscribed data. The pseudo-labels can either be one-best transcripts or decipherment lattices. Unlike conventional SST, we have no seed model for the target language, since there is no equivalence between the outputs of the phone recogniser and the target language graphemes.  We therefore choose to train a model with flat-start lattice-free MMI (LF-MMI)  \cite{hadian2018flat}, initialising the lower layers of the model with the universal phone recogniser, but using a randomly-initialised output layer.

\section{Experiments}
\label{sec:experiments}

To demonstrate that decipherment can be used for cross-lingual transfer, we trained an universal phone recogniser on English, French, German, Spanish, Russian and Polish and we performed decipherment experiments on Bulgarian (BUL), Czech (CES), Hausa (HAU), Portuguese (POR), Swahili (SWA), Swedish (SWE) and Ukrainian (UKR).

\subsection{Setup}

\begin{table*}[t]

\centering

\caption{Word Error Rate (WER) of ASR models trained for Bulgarian (BUL), Czech (CES), Hausa (HAU), Portuguese (POR), Swahili (SWA), Swedish (SWE) and Ukrainian (UKR).}

\begin{tabularx}{\textwidth}{Xccccccc}
                    & \textbf{BUL} & \textbf{CES} & \textbf{HAU} & \textbf{POR} & \textbf{SWA} & \textbf{SWE} & \textbf{UKR}\\

\hline
\textbf{Oracle with GlobalPhone LM}   & 8.5  & 11.7  & 7.6  & 15.0  & N/A  & 16.8  & N/A \\
\textbf{Oracle with CommonCrawl LM}   & 7.9  & 12.9  & 12.2  & 17.0  & 10.2  & 23.3  & 8.4 \\
\hline
\textbf{Phone-mapping}                & 35.0  & 44.3  & 43.4  & 52.8  & 82.5  & 66.8  & 35.7 \\
\textbf{ + semi-supervised training}  & 12.6  & 16.8  & 25.3  & 21.8  & 52.8  & 37.5  & 12.1 \\
\hline
\textbf{Decipherment}                 & 31.0  & 49.0  & 70.8  & 53.8  & 105.3  & 93.5  & 34.5 \\
\textbf{ + semi-supervised training}  & 10.7  & 16.5  & 30.3  & 21.2  & 98.3  & 55.8  & 10.3 \\

\end{tabularx}

\label{tab:decipherment}

\end{table*}

Our experiments were performed using the GlobalPhone corpus~\cite{schultz2013globalphone}. This corpus comes with data from various languages and contains lexicons, which can be mapped to X-SAMPA, enabling the pooling of phonemes across languages. All these properties make GlobalPhone an ideal test-bed for evaluation of zero-resource cross-lingual transfer with decipherment.

To train the \textbf{Universal Phone Recogniser} as a multilingual model with a shared phone-set we pooled 20 hours of English LibriSpeech ~\cite{panayotov2015librispeech} with the training data from GlobalPhone German, French, Spanish, Russian and Polish~\cite{schultz2013globalphone}, 110 hours of data in total. The multilingual model was a small time-delayed neural network~\cite{povey2018semi} with 18 hidden layers each having 798 hidden layer size and 90 bottleneck size. In total the model had 7.2M parameters, used 40 dimensional cepstral mean and variance normalised MFCC features as inputs, and was trained with LF-MMI~\cite{povey2016purely} using the Kaldi toolkit~\cite{povey2012kaldi}. We used a phone-bigram language model estimated on the multilingual training data for cross-lingual phone decoding.

We trained all \textbf{Language Models} on the text data from CommonCrawl~\cite{buck2014commoncrawl}. Because the CommonCrawl text data is noisy we preprocessed it as follows. We performed word tokenization and removed tokens consisting only of non-alpha-numeric characters. We mapped words containing characters outside of the target language alphabet or containing letters repeated at least 3 times in a row to \texttt{<unk>} and removed sentences containing any word longer than 20 characters or with three consecutive single-letter words.
We trained language models with SRILM~\cite{stolcke2002srilm} on up to 1B tokens and we pruned the language models to only contain 300k most frequent words. We did not use the pretrained GlobalPhone language models because we found that some of them had been also trained on the training transcripts which could bias the results of semi-supervised training. But for completeness we also include the oracle results obtained with GlobalPhone LMs in Table~\ref{tab:decipherment}.

The \textbf{Decipherment Model} was trained on 20 minutes of the shortest utterances from the development set of each language.  We increased the power of the character-level language model over successive epochs from a bigram up to a 5-gram, performing 20 iterations of full-batch training with each language model. These grapheme language models were trained on the first 50k lines of the normalised CommonCrawl text data.
To prevent issues with bad initialisation of the decipherment model we performed 50 random restarts with the bigram language model and we picked the model with the best likelihood on the training data for successive training~\cite{berg2013decipherment}. To speed up training with the larger grapheme models, we pruned the lexical model to retain probabilities for only the top 20 phones for each grapheme after training with the bigram grapheme language model. After this stage we smoothed the lexical model and continued training with the CommonCrawl word language model. Finally, we smoothed the lexical model again and deciphered the GlobalPhone training data.
Note that during training we used a word language model containing only 100k most frequent words but during inference we used the language model containing 300k most frequent words.

Subsequently we performed two iterations of \textbf{Semi-Supervised Training} with the deciphered lattices representing alternative pseudo-labels.
In the first iteration we used these lattices for flat-start LF-MMI training~\cite{hadian2018flat}. Instead of training the model from scratch, we replaced the output layer of the multilingual model with a new layer producing pseudo-likelihoods for mono-graphemes.
In the second iteration we used the mono-grapheme model to re-decode the training data. To prevent overfitting to the training data we did not continue training the mono-grapheme model, but again replaced the output layer of the pretrained multilingual model and used that model for training. This time we used bi-grapheme targets, because of now having more reliable pseudo-labels with which to estimate the state clustering tree. Since the decipherment tends to produce a lot of deletion errors we used a deletion penalty during decoding to allow the model to learn to fix them~\cite{fainberg2019lattice}.

We compared the performance with two other approaches. The first was standard supervised training, called Oracle in Table~\ref{tab:decipherment}, and in the second we used linguistic knowledge to map phones from the target language to the closest phone in the pooled multilingual phone set to generate pseudo-labels for semi-supervised training~\cite{vu2011cross,prasad2019building}, called Phone-mapping in Table~\ref{tab:decipherment}. In both approaches we also initialised the models by replacing the output layer of the pretrained multilingual model.

\subsection{Results}

Our results in Table~\ref{tab:decipherment} show that decipherment achieves comparable or better results to the hand-crafted phone-mapping approach for Bulgarian, Czech, Portuguese and Ukrainian, which are well-resourced languages. For all these languages, decipherment followed by semi-supervised training is only 2 -- 4\% absolute worse than Oracle with CommonCrawl LM. Swedish is the only well-resource language for which decipherment performs much worse with the absolute difference of 32.5\%.

For lower-resourced languages Swahili and Hausa, phone mapping achieves better results. For Swahili we were unable to achieve a successful decipherment. By listening to the GlobalPhone Swahili data we identified several problems, including beeps at the beginning of each utterance and a lot of leading and trailing silence. Even when we removed the beeps and leading and trailing silence in the Swahili data the performance was bad (as reported in Table~\ref{tab:decipherment}). Therefore, we decided to evaluate the method also on Swahili data from the ALFFA corpus~\cite{gelas2012developments} which has been used for unsupervised speech recognition experiments in~\cite{baevski2021unsupervised}. On this dataset decipherment followed by semi-supervised training achieves a WER of 41.8\% which compares to 32.2\% reported in~\cite{baevski2021unsupervised} and 24.6\% achieved by our oracle model. These Swahili results suggest that in order to be able to decipher speech from a new language we need to find speech amenable to decipherment. 

We believe that our results in all languages could be further improved by using a better universal phone recogniser~\cite{li2020universal}, a better pre-trained model for initialisation~\cite{baevski2020wav2vec} and by leveraging more crawled data for SST~\cite{wallington2021learning,carmantini2019untranscribed}. Our results are in-line with our previous work on conventional low-resource SST where we showed that it is possible to perform SST even with a bad seed acoustic model provided that we have a good language model~\cite{wallington2021learning}.

\section{Conclusions and Future Work}
\label{sec:conclusions}

We presented a method for zero-resource cross-lingual transfer of ASR models based on decipherment that allows training of ASR models using only untranscribed speech, text corpora and a universal phone recogniser.  Across seven test languages our method was able to produce a working acoustic model for six, which could be further improved by using more untranscribed data for SST.
In future we plan to apply decipherment to more challenging languages, but we believe that for this it may be necessary to train a more robust universal phone recogniser that works reliably across a wider range of various languages. %
We also intend to improve our decipherment algorithm to enable selection of utterances that can be reliably deciphered.  Further, we hope to replace the universal phone recogniser with automatic unit discovery to create a pronunciation lexicon-free alternative for unsupervised speech recognition~\cite{baevski2021unsupervised}.

\bibliographystyle{IEEEtran}

\bibliography{mybib}

\end{document}